%% file: main.tex
\documentclass{sig-alternate-05-2015}
\usepackage[sort,numbers]{natbib}

\usepackage{hyperref}
\hypersetup{colorlinks=true, urlcolor=blue, citecolor=cyan, pdfborder={0 0 0},}
\usepackage{soul}
\usepackage{natbib}
\usepackage{authblk}

\title{Distributed Representations for Biological Sequence Analysis}

\author[*]{Dhananjay Kimothi}
\author[!]{Akshay Soni}
\author[*]{Pravesh Biyani}
\author[\dagger]{James M. Hogan}

\affil[*]{IIIT Delhi, India}
\affil[!]{Yahoo! Research, Sunnyvale CA, USA}
\affil[\dagger]{Queensland University of Technology (QUT), Australia}

\begin{document}

\maketitle
\input{abstract.tex}
\input{introduction.tex}
\input{method.tex}

\input{experiments.tex}

\input{conclusions.tex}
\bibliographystyle{acm}
\bibliography{main}  
\end{document}

%% file: abstract.tex
\begin{abstract}
Biological sequence comparison is a key step in inferring the relatedness of various organisms and the functional similarity of their components. Thanks to the Next Generation Sequencing efforts, an abundance of sequence data is now available to be processed for a range of bioinformatics applications. Embedding a biological sequence -- over a nucleotide or amino acid alphabet -- in a lower dimensional vector space makes the data more amenable for use by current machine learning tools, provided the quality of embedding is high and it captures the most meaningful information of the original sequences. 

Motivated by recent advances in the text document embedding literature, we present a new method, called {\tt seq2vec}, to represent a complete biological sequence in an Euclidean space. The new representation has the potential to capture the contextual information of the original sequence necessary for sequence comparison tasks. We test our embeddings with protein sequence classification and retrieval tasks and demonstrate encouraging outcomes. 
\end{abstract}

%% file: introduction.tex
\section{Introduction}
Nucleic acids and proteins are frequently modeled through their primary structure or {\em sequence}, a character string representation based on their constituent sub-units, with each nucleotide or amino acid base drawn from an appropriate alphabet. While biological sequences are necessarily an abstraction from the complex three-dimensional structure of the original molecule, the representation retains information sufficient to infer a good deal about the nature of the organism and the function of its structural components. By analogy, tailored comparison of different biological sequences or sub-sequences may allow the computational biologist to infer the functional similarity or relatedness of the organisms or their components.  

General purpose sequence comparison has thus received considerable attention in the bioinformatics literature, with many approaches based on alignment of a pair of sequences, whether in their entirety (\emph{Global Alignment}, introduced by Needleman and Wunsch~\cite{Needleman1970}) or by considering local regions of high similarity (\emph{Local Alignment}, due to Smith and Waterman~\cite{Smith1981}), with distances weighted according to their biological effect via substitution matrices such as BLOSUM~\cite{henikoff1992amino} for protein sequences. These approaches define a formal metric between sequences and may be regarded in principle as a canonical distance measure, but their direct application is limited by efficiency considerations. Each algorithm is based on dynamic programming and has computational complexity quadratic in the length of the sequences. For large sequences such an approach can be very slow, specially for applications which require many such comparisons. These limitations led naturally to the development of heuristic methods for searching large sequence databases -- the most prominent of which is BLAST, The Basic Local Alignment Search Tool~\cite{Altschul1990} -- and more efficient algorithms for alignment of large numbers of short sequences, such as CLUSTALW~\cite{thompson2002multiple}. Probabilistic alternatives, such as the profile Hidden Markov Model (HMMER)~\cite{eddy2007hmmer}, have enjoyed considerable success for protein database search and sequence alignment, but are less widely used than the pairwise methods described above.

The explosion in the availability of biological sequence data arising from Next Generation Sequencing (see Metzker~\cite{MetzkerNGS} for a review) has driven work toward more scalable approaches to sequence comparison, notably through so-called \emph{alignment-free} methods (see Song et.al.~\cite{Song2014b}). Such approaches avoid the costly reliance on dynamic programming in favor of faster alternatives, usually based on tokenization of the sequence into a set of substrings or {\tt kmers}, words of length $k$ extracted from the sequence~\cite{mizuta2014new,yu2010novel,melko2004distribution,leimeister2014fast,wen2014k}. 

These methods are based on two key steps. First, a lexical representation for the sequence is obtained by extracting the underlying kmers, usually by  traversing a window of size $k$ across the original sequence and recording those which appear. Note that this is normally an overlapping set, so that {\tt ACGTTA} will yield {\tt ACG, CGT, GTT} and {\tt TTA} for $k=3$, though some authors work with a reduced lexicon. Given this kmer set, the second step is to find an alternative representation in a new domain (like $\mathbb{R}^n$) where a formal distance metric is defined. The main goal is that the representation should be obtained in a manner such that the pairwise similarity of any two biological sequences is approximately preserved. In other words, biological sequences which are functionally or evolutionarily similar should exhibit a relatively small separation in the new space. The main advantage of these transformations is the flexibility to apply various machine learning algorithms within the new feature space. Thus the efficacy of this new representation depends on the appropriate application of ML based tools, which in turn depends on the specific bioinformatics application at hand. \par 

This paper introduces a novel representation methodology for biological sequence comparison and classification. Although the proposed technique may be applied to sequences over an arbitrary alphabet, in this work we explore the effectiveness of this embedding in the context of protein sequence classification. Before presenting a more detailed view of our contribution, we consider the protein classification problem in more depth. \par 

\begin{figure*}[t]
\centering
\begin{tabular}{cccc}
\includegraphics[width=1.4in]{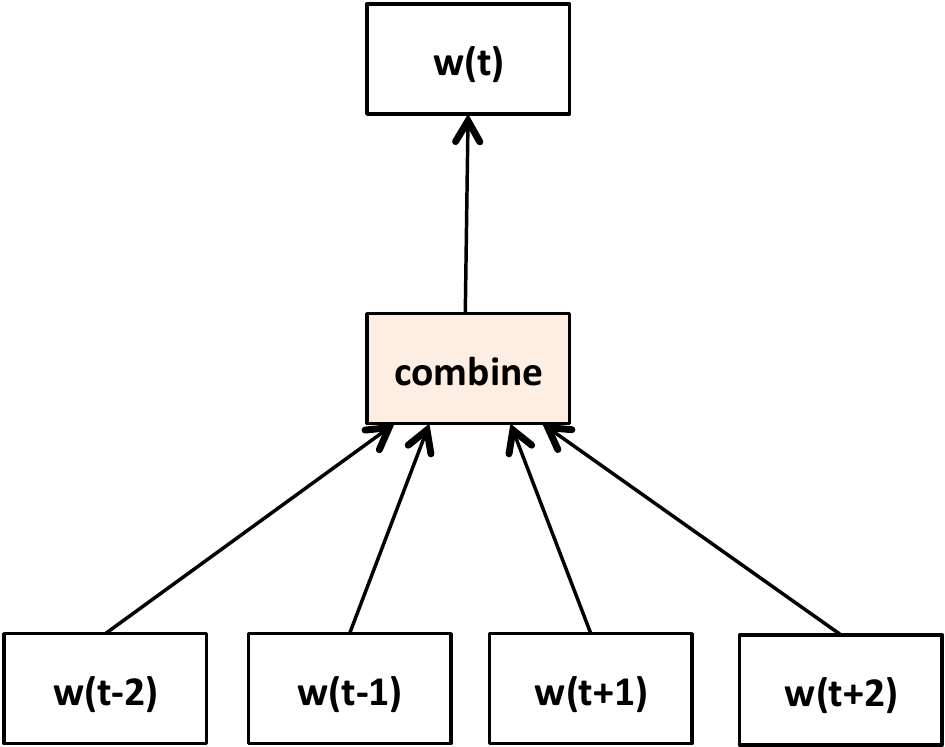} & \includegraphics[width=1.4in]{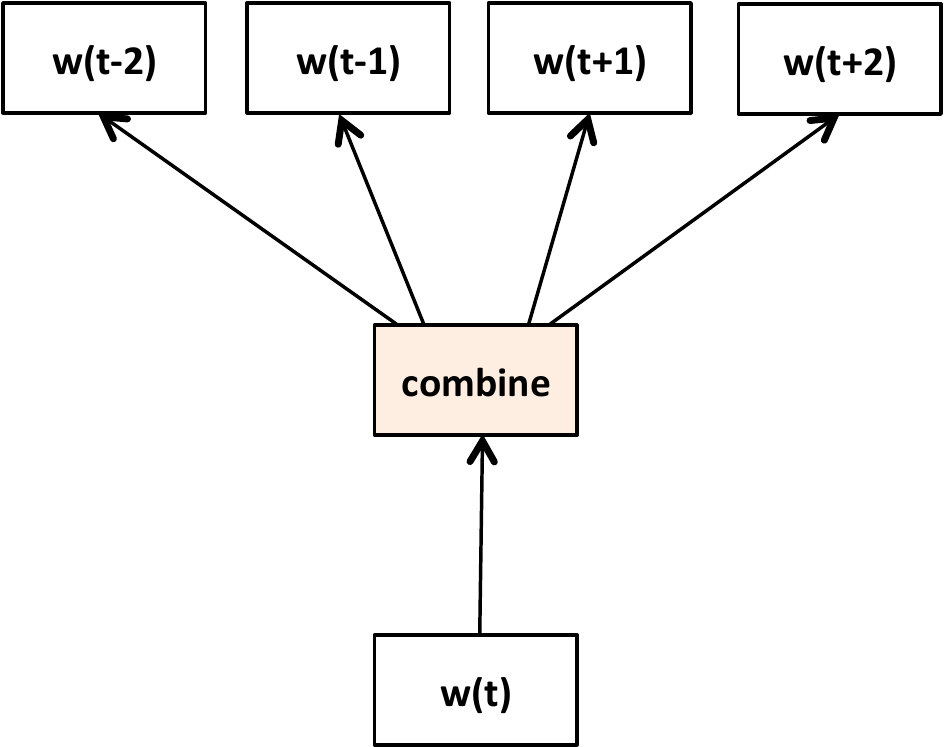} & 
\includegraphics[width=1.7in]{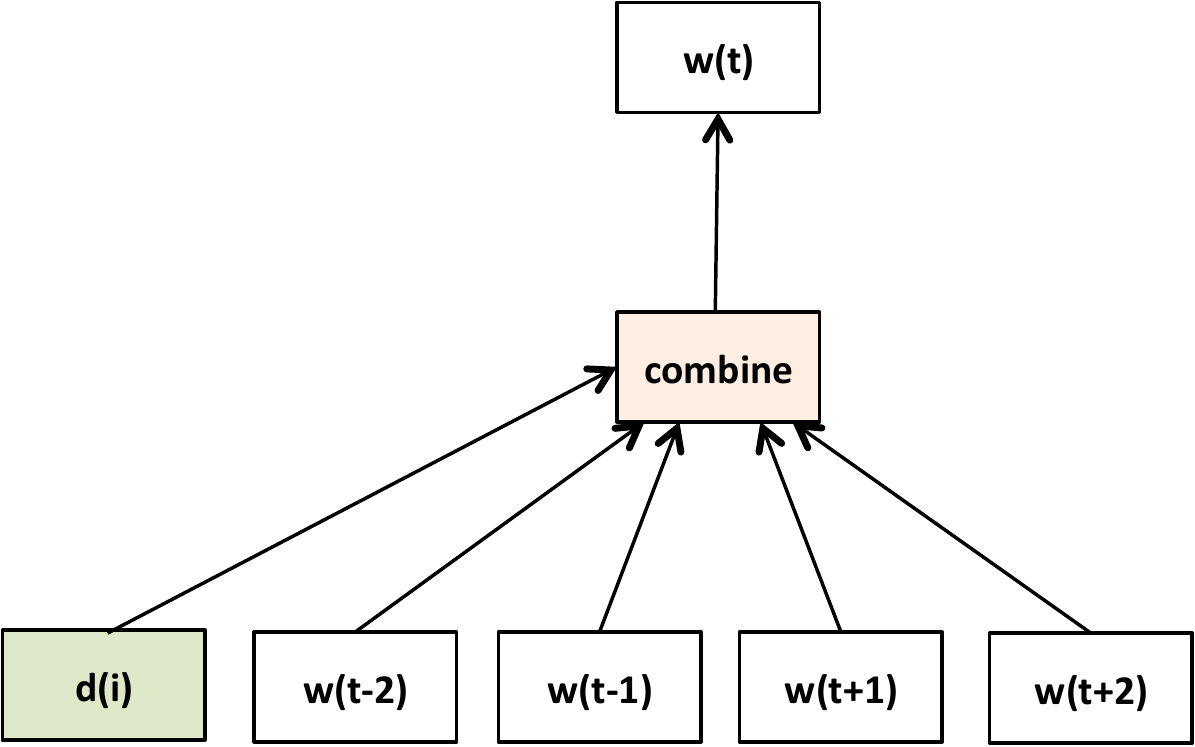} & \includegraphics[width=1.7in]{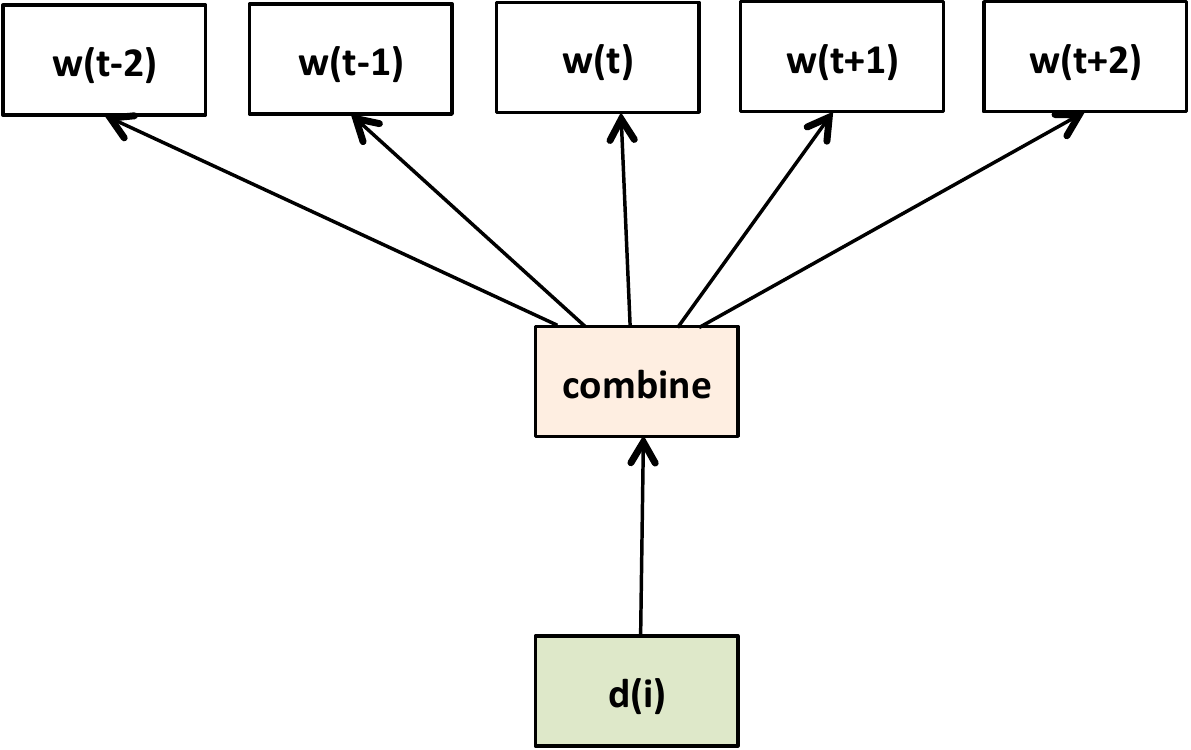}\\
{\tt word2vec}: CBOW & {\tt word2vec}: skip-gram & {\tt doc2vec}: DM & {\tt doc2vec}: DBOW
\end{tabular}
\caption{Architecture for CBOW and skip-gram for {\tt word2vec}, DM and DBOW for {\tt doc2vec}. Notice the similarity between CBOW and DM, skip-gram and DBOW.}
\label{figure:architectures}
\end{figure*}

\subsection{Protein Families}
A protein family is a group of related proteins exhibiting significant structural similarity at both the sequence and molecular level. Large scale assignment of proteins to families grew out of the work of Dayhoff~\cite{dayhoff1974computer}. Sequence homology is regarded as a direct reflection of evolutionary relatedness, and members of the same protein family may also exhibit similar secondary structure through common functional units called~\emph{domains}. Domains present an additional organizing principle, allowing protein families to be grouped into super  families~\cite{dayhoff1975evolution} on the basis of their domains, even if sequence homology is limited for some pairs of protein entries. These families and super-families are documented in detail in databases such as SCOP~\cite{conte2000scop} and their successors. 

Direct determination of 3D protein structure is difficult and time consuming, requiring complex experimental techniques such as X-ray crystallography or NMR spectroscopy. For this reason experimental confirmation is available for only a small fraction of the proteins available, leading researchers to develop methods~\cite{Enright2002,Bork1998,Remmert2011} which use only primary structure for family and thus functional identification.

\subsection{Related Work}
The problem of protein sequence labeling has a long history, and a mix of generative, discriminative, probabilistic and similarity-based clustering approaches has been employed in past. Early approaches are well summarized in  Leslie et al.\cite{leslie2004mismatch}; many of these works, including~\cite{Cai2003,Huynen2000} and that of the Leslie group adopt a discriminative approach via the support vector machine. While some of these studies include feature sets based on physical properties such as hydrophobicity, van der Waal volume, polarity and surface charge, Leslie et.al.\cite{leslie2004mismatch}  work directly with the properties of the sequence. This approach underpins the present work. 

\subsubsection{Distributed Representation of Sequences}
Recent work in the Natural Language Processing community (Mikolov et.al.~\cite{Mikolov2013word2vec} and Le and Mikolov \cite{le2014distributed}) has highlighted the value of \emph{word embeddings} in the identification of terms with a similar linguistic context. In this approach, known as {\tt word2vec}, words or phrases from the lexicon are mapped to vectors of real numbers in a low-dimensional space. By training a (shallow) Neural Network over a large text corpus, words with similar linguistic context may be mapped close by in the Euclidean space. \par 
 
 Asgari and Mofrad \cite{Asgari2015} recently applied the {\tt word2vec} framework to the representation of, and feature extraction from, biological sequences. The embeddings generated from their algorithm is named {\tt BioVec} for general biological sequences and {\tt ProtVec} for the specific case of proteins. The representations {\tt BioVec} and {\tt ProtVec} may be utilized for several common bioinformatic tasks. Here, a biological sequence is treated like a sentence in a text corpus while the kmers derived from the sequence are treated like words, and are given as input to the embedding algorithm. The authors in \cite{Asgari2015} used the representations thus obtained for the protein classification task, reporting an accuracy of $93\%\pm 0.06$ for $7027$ protein families. \par 
 
 \subsection{Motivation and Contribution}
 Apart from this encouraging performance on the protein classification task, a word embeddings based approach offers other advantages. First, the size of the data vector obtained by using embeddings is much smaller in comparison to the original sequence, thus making the vectors obtained amenable for various machine learning algorithms. Indeed, one needs to compute the representation only once, while the machine learning tasks on the representations obtained are relatively inexpensive computationally, making this technique suitable for large scale bioinformatics applications. The efficacy of this approach is heavily dependent on the quality of the embeddings obtained, which in turn crucially depend on various factors like context size, vector size and finally the training algorithm. While Asgari and Mofrad gave a promising initiation towards the embeddings approach, the interplay of these factors is not well understood. It should be noted that  Asgari and Mofrad used the representation obtained for each of the biological words or kmers in order to derive the representation for the whole biological sequence. One potential weakness of such an approach is that it does not {\it fully} capture the order in which the words occur in the original sequence. \par 
 In this paper, we estimate the representation of a complete protein sequence as opposed to obtaining those for the individual kmers. Our algorithm is based on the {\tt doc2vec} approach, which is an extension of the original {\tt word2vec} algorithm. We call our proposed framework {\tt seq2vec} and use it to obtain protein sequence embeddings. 
 Sequences obtained by {\tt seq2vec} compares favorably with {\tt ProtVecs} on different classification  tasks and is discussed in section \ref{sec: exp}. For evaluation purposes we apply {\tt seq2vec} to the problem of learning  vectors for the set of protein sequences provided by~\cite{Asgari2015}. We then use these vectors as the basis for the protein classification task, following the approach of~\cite{Asgari2015} for performance evaluation of {\tt ProtVec}s. We also use these vectors along with {\tt ProtVec}s for classification of sequences into their respective families based on the Euclidean neighborhood of the sequence. The classification accuracy thus obtained is used to infer the  quality of the embedding, with higher accuracy suggesting a better quality embedding. Further, to examine performance relative to alignment based methods, we compare these results with sequence classification based on BLAST results. Since BLAST is a retrieval tool, we use retrieval results obtained for a query sequence to predict its class.  A class is assigned to the query based on the class of each of the top-N retrieved sequences according to a majority voting scheme. 

%% file: method.tex
\newpage
\section{Our Approach}\label{method}
Since our approach is based on text embedding ideas from the NLP literature, we start with a brief review of {\tt word2vec} and {\tt doc2vec}, followed by details of {\tt seq2vec}, which uses {\tt doc2vec} at its core to learn embeddings for biological sequences. 

\subsection{Distributed Representations for Text }\label{subsec: Distrb rep in NLP}
Learning representative embeddings of words and documents is a fundamental task in many NLP and machine learning applications. While simple bag-of-words based vector representation of documents have been used successfully in tasks like spam-email detection, they suffer from high dimensionality and an inability to capture complex information such as semantics and context of the underlying text. This makes their use limited for advanced tasks like multi-label learning, humor detection, hate detection, etc. 

To alleviate the above mentioned shortcomings, {\tt word2vec} \cite{Mikolov2013word2vec} models -- continuous bag-of-words (CBOW) and skip-gram --  were recently proposed. These models are shallow, two-layer neural networks that are trained to reconstruct linguistic contexts of words. Given a corpus of text, these models assign a vector of specified dimension to each word such that words that share common contexts in the training corpus are located close to each other in the shared embedded space. {\tt word2vec} provides two architecture choices -- CBOW and skip-gram. With CBOW, the model predicts the current word by using a few surrounding context words. On the other hand, skip-gram uses the current word to predict the surrounding context words. CBOW is generally faster and is the preferred choice when a large corpus is available for training. Skip-gram is used when training data is small and it also provides better representations for infrequent words. The vectors learned from these models are powerful in terms of their ability to capture semantics i.e. words with similar meanings are embedded close to each other -- for example, the word ``strong"  and ``powerful" would be closer to each other than ``strong" and ``London" in the Euclidean space. 

Follow-up research extended {\tt word2vec} models to embed sentences and paragraphs --  more generally, documents --  in a vector space \cite{ grefenstette2013multi, mikolov2013distributed}. The simplest approach is to use the weighted sum of word-vectors learned using {\tt word2vec}. This is conceptually simple but does not take the order of words into consideration, which implies that sentences composed of the same words, but in a different order, will be represented by same vectors. To address these issues, {\tt doc2vec} \cite{le2014distributed} models -- distributed-memory (DM) and distributed bag-of-words (DBOW) -- have been proposed. DM is akin to the CBOW architecture in {\tt word2vec} and learns a vector representation of each word in the corpus of documents, along with the vectors for documents as well. This is done by predicting the current word using the context words and the document vector. DBOW is similar to the skip-gram architecture in {\tt word2vec} and uses the document vector to predict the context words. DBOW is conceptually simple and does not require storage of the word vectors if the eventual task is to learn the document vectors. Note that in {\tt doc2vec}, the word vectors are global and are updated for each context, while document vectors are local and are only updated for contexts from this document.

An overall architecture diagram for CBOW, skip-gram, DM, and DBOW is shown in Figure \ref{figure:architectures}.

\subsection{seq2vec}\label{subsec:seq2vec}
In order to port the advantages of using {\tt doc2vec} in place of combining word vectors learned using {\tt word2vec}, we improve upon {\tt BioVec} \cite{Asgari2015}, which is a {\tt word2vec} based approach, by using the {\tt doc2vec} framework instead. We call our approach {\tt seq2vec}, a method to embed an arbitrary biological sequence into a vector space.

Unlike a document, which may be seen as an array of words with a certain linguistic structure, biological sequences are strings of letters selected from an alphabet -- for DNA sequences, $\{$A,~C,~G,~T$\}$, the set of nucleotides Adenine, Cytosine, Guanine and Thymine, while for protein sequences the alphabet consists of $20$ letters, each representing an amino acid. Since there is no clear notion of words in biological sequences, we need to break these sequences during the preprocessing stage to represent them as an array of kmers -- a unit made of $k$ consecutive letters. 

We employ two ways of processing the sequences -- non-overlapping and overlapping processing. In non-overlapping processing, we generate $k$ new sequences from a given sequence by shifting the starting point from where we start constructing the kmers. For example, given a hypothetical sequence
\begin{equation*}
\text{QWERTYQWERTY}
\end{equation*}
and $k = 3$, we generate three new sequences as follows:
\begin{eqnarray*}
&& \text{Seq 1: QWE~RTY~QWE~RTY} \\
&& \text{Seq 2: WER~TYQ~WER} \\
&& \text{Seq 3: ERT~YQW~ERT}.
\end{eqnarray*}
In overlapping processing, we generate one sequence by breaking the original sequence into overlapping kmers. While the overlap itself is a parameter, in this paper we focus on kmers shifted by one letter at a time. For instance, overlapping processing of the hypothetical sequence above yields:
\begin{equation*}
\text{QWE~WER~ERT~RTY~TYQ~YQW~QWE~WER~ERT~RTY.}
\end{equation*}

After this preprocessing, each sequence is treated as a document with kmers as corresponding words. We then apply the standard {\tt doc2vec} model to learn the vector representation of each sequence in the corpus. The choice of using either DM, or DBOW, or both (as in combining representations learned from DM and DBOW) depends on the underlying task, and needs to be made using cross-validation. Similarly, other learning hyper-parameters like context window size, vector space dimension, hierarchical softmax and/or negative sampling for learning procedure, and the threshold for sub-sampling high-frequency words are all selected using cross-validation.

%% file: experiments.tex
\section{Experimental Results}\label{sec: exp} 
\begin{figure}[t]
    \centering
        \includegraphics[width=0.4\textwidth]{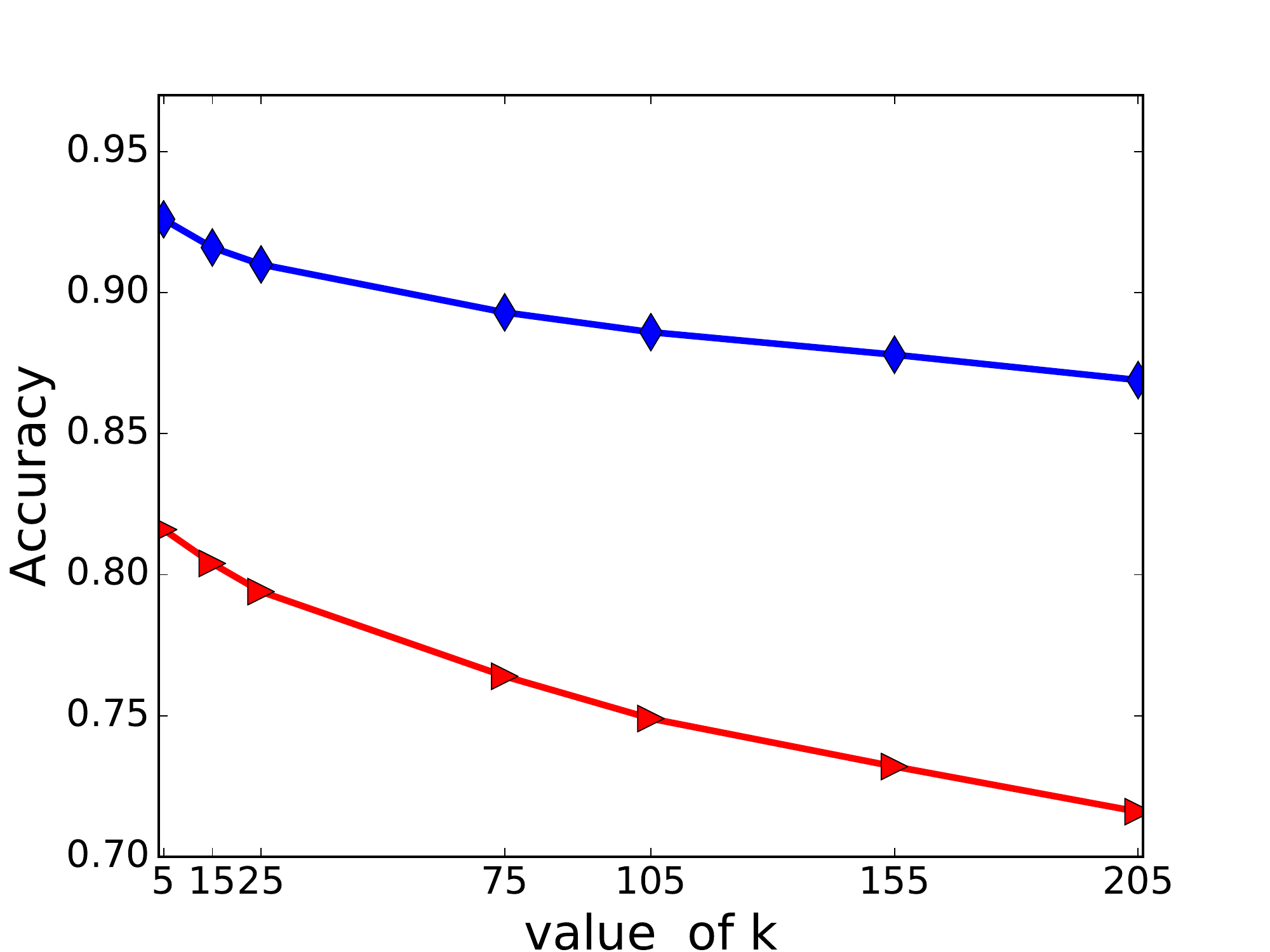}
    \caption{Performance of \textcolor{blue}{non-overlapping} (\textcolor{blue}{--$\blacklozenge$--$\blacklozenge$--}) vs \textcolor{red}{overlapping} (\textcolor{red}{--$\blacktriangleright$--$\blacktriangleright$--}) pre-processing for the task of kNN based classification for different number of nearest neighbors. This is with best choice of other parameters -- dimension: $250$, {\tt kmer} size: $3$, context size: $5$ ($25$ for overlapping). It is clear that non-overlapping processing performs significantly better than the overlapping processing.}
    \label{fig:ovr-non-ovr}
\end{figure}

\begin{table*}[ht]
\centering
\begin{tabular}{c|c||c|c|c|}
\hline
\multicolumn{1}{|l|}{} & \textbf{\# of classes} & \textbf{Specificity (\%) } & \textbf{Sensitivity (\%) } & \textbf{ Accuracy (\%)}\\ 
\hline \hline
\multicolumn{1}{|l|}{{\tt seq2vec}} & \multicolumn{1}{c||}{\begin{tabular}[c]{@{}l@{}}1000\\ 2000\\ 3000\end{tabular}} & \multicolumn{1}{c|}{\begin{tabular}[c]{@{}l@{}}\textbf{97.49 $\pm$ 0.05}\\ \textbf{97.01 $\pm$ 0.04}\\ \textbf{96.67 $\pm$ 0.05}\end{tabular}} & \multicolumn{1}{c|}{\begin{tabular}[c]{@{}l@{}}92.72 $\pm$ 0.06\\ 93.07 $\pm$ 0.07\\ 93.18 $\pm$ 0.08\end{tabular}}  & \multicolumn{1}{c|}{\textbf{\begin{tabular}[c]{@{}l@{}}95.10 $\pm$ 0.05\\ 95.04 $\pm$ 0.05\\  94.92 $\pm$ 0.06\end{tabular}}} \\ \hline
\multicolumn{1}{|l|}{{\tt ProtVec}} & \multicolumn{1}{c||}{\begin{tabular}[c]{@{}l@{}}1000\\ 2000\\ 3000\end{tabular}} & \multicolumn{1}{c|}{\begin{tabular}[c]{@{}l@{}}91.61 $\pm$ 0.05\\ 90.10 $\pm$ 0.07\\88.61 $\pm$ 0.09\end{tabular}}   & \multicolumn{1}{c|}{\begin{tabular}[c]{@{}l@{}}\textbf{95.40 $\pm$ 0.04}\\ \textbf{95.75 $\pm$ 0.05}\\ \textbf{95.95 $\pm$ 0.05}\end{tabular}} & \multicolumn{1}{c|}{\begin{tabular}[c]{@{}l@{}}93.50 $\pm$ 0.05\\ 92.92 $\pm$ 0.05\\ 92.27 $\pm$ 0.06\end{tabular}}\\ \hline
\end{tabular}
\caption{Binary Classification: performance of {\tt seq2vec} and {\tt ProtVecs} for top 1000, 2000 and 3000 protein families.}
\label{binary-clf}
\end{table*}

In order to demonstrate the advantages of {\tt seq2vec} over {\tt BioVec}, we carefully designed our experiments to compare the two sets of embeddings on multiple fronts. First, we use the protein vectors learned using {\tt seq2vec} and {\tt ProtVecs} as features and compare them for the task of protein classification using SVMs. This involves experiments with both binary and multiclass classification problems. Next, since distributed representations embed similar  sequences in proximity to each other, we use $k$-nearest neighbors (kNN) to retrieve the $k$ nearest sequences in the vector space and see how successful are we in predicting the family of a test sequence based on a majority vote. Finally, we compare this retrieval task with BLAST, the standard approach for rapid determination of sequence similarity, and notice that at present BLAST performs significantly better than the embedding based approaches. This motivates our discussion on future work in the next section. 

We employ {\tt seq2vec} for learning distributed vector representations of $324018$ protein sequences from \cite{Asgari2015} which are originally taken from the Swiss-Prot database \cite{uniprot2014uniprot}, they also provided a meta file which contains the family labels along with other description of protein sequences. We extracted the family label information from this meta file. To our surprise we found that the unique family labels extracted from the meta file are $6097$, which is different than the total number of families($7027$) reported by \cite{Asgari2015}. On further probing we found that in classification result file provided by \cite{Asgari2015} 
there are multiple entries for some families but the number of samples  corresponding to such  entries are different. Adding up the number of sequences for such entries for a given family matches to the actual number of sequences as per the meta file, which suggests that the samples of few families are splitted into sub groups by \cite{Asgari2015} in their experimental setup, which is the reason for difference in number of families reported in this work and \cite{Asgari2015}. In our experiments we utilized the  labels from meta file and based on it categorized sequences into   $6097$ families, out of these, $3861$ families have fewer than $10$ sequences, $2472$ families have number of sequences in the range $11$ - $100$, $609$ families have number of sequences in the range $101$ - $1000$, and remaining $25$ families contain more than $1000$ sequences. 

In order to choose a good set of hyper-parameters -- including context size, vector space dimension, non-overlapping vs. overlapping pre-processing, model architecture -- we did a search over a range of these parameters for the task of kNN based classification and chose parameters which gave the best accuracy for sequence retrieval from the same family as the query sequence on a separate validation set. We did this experiment using sequences from the $25$ biggest families having more than $1000$ sequences. Overall, from this experiment we chose dimension: $250$, kmer size: $3$, context size: $5$, pre-processing: non-overlapping, and  architecture: DM (see Figure.\ref{fig:ovr-non-ovr}).All the experiments were performed using Gensim \cite{rehurek_lrec}
  
\begin{table*}[ht]
\centering
\begin{tabular}{|c||c|c|c|}
\hline
        & \textbf{Precision(\%)}    & \textbf{Sensitivity (\%) }   & \textbf{ Accuracy (\%)} \\ \hline \hline
{\tt seq2vec} & \textbf{83.37 $\pm$ 0.052} & \textbf{81.69 $\pm$ 0.067}  & \textbf{81.29 $\pm$ 0.057} \\ \hline
{\tt ProtVec}  & 79.01 $\pm$ 0.071  & 76.78 $\pm$ 0.082 & 76.70 $\pm$ 0.080 \\ \hline
\end{tabular}
\caption{Multiclass Classification: Performance of {\tt seq2vec} and {\tt ProtVec} for classification of sequences from top 25 families.}
\label{tble:multi-clf}
\end{table*}

\subsection{Protein Classification}
As discussed above we first compare representations learned from {\tt seq2vec} and {\tt BioVec} for the task of binary and multiclass classification on the following three metrics:

\begin{eqnarray*}
{\rm Specificity} &=&  \frac{{\rm TN}}{{\rm TN+FP}}\\
{\rm Sensitivity} &=&  \frac{{\rm TP}}{{\rm TP+FN}}\\
{\rm Accuracy} &=&  \frac{{\rm TN}+{\rm TP}}{{\rm TN+TP+FN+FP}}
\end{eqnarray*}
where TN is true-negatives, TP is true-positives, FP is false-positives, and FN is false-negatives. Specificity (true negative rate) measures the proportion of negatives that are correctly classified. Sensitivity (true positive rate or recall) measures the proportion of positives that are correctly classified. Accuracy measures the proportion of examples that are correctly classified. Depending on the application, the relative importance of these metrics may change.

Note that we use the vectors provided by authors of \cite{Asgari2015} for doing experiments involving {\tt ProtVec}\footnote{http://llp.berkeley.edu/}.

\subsubsection{Binary Classification}\label{binary clf section}
This experiment exactly follows the setting used by \cite{Asgari2015}. The task is to distinguish sequences of a given family from all other families. For each family, the corresponding sequences constitute the positive class, and the same number of sequences, randomly chosen from rest of the families, forms the negative class. A binary classifier is trained and the metrics are computed using $10$-fold cross validation.

While the results reported in \cite{Asgari2015} also trained classifiers for families with fewer than $10$ sequences, we included families with at least $10$ sequences in order to perform a proper $10$-fold validation. Also, more data makes the learned classifiers more stable and the reported results are stable and reproducible. 

The exact choice of SVM hyper-parameters, including the type of kernel and $C$, are not clearly mentioned in \cite{Asgari2015}. We chose to use an SVM classifier with linear kernel and C equal to $1.0$, these choices being made based on grid search. We did three set of experiments with the $1000$, $2000$, and $3000$ biggest families. The results reported in Table \ref{binary-clf} show that the accuracy and specificity obtained using {\tt seq2vec} is consistently better than {\tt ProtVec} while {\tt ProtVec} performs slightly better in terms of sensitivity than {\tt seq2vec}.

\subsubsection{Multiclass Classification}
The very high values for metrics as reported in Table \ref{binary-clf} for the binary classification task is possibly due to the way in which the negative class is constructed -- by randomly selecting sequences from all the remaining families. Since the number of possible negative sequences is very large, the randomly selected sequences would generally be far away from the positive sequences, making the positive and negative classes far apart, resulting in an easy classification problem. 

To alleviate this issue with the experimental setup of \cite{Asgari2015}, we adopt a different approach. In order to compare {\tt seq2vec} with {\tt ProtVecs}, we do a multiclass classification to classify the sequences into their corresponding families. We did this experiment using the $25$ biggest families consisting of more than $1000$ sequences each. Then we trained a multiclass SVM classifier with a linear kernel using the one-vs-rest strategy. We set $C = 1$ for {\tt seq2vec} and $C = 7.5$ for {\tt ProtVecs} based on grid search. The evaluation metrics are then computed using $10$-fold cross validation and the results are reported in Table \ref{tble:multi-clf}.

As shown in Table \ref{tble:multi-clf}, {\tt seq2vec} perform much better than {\tt ProtVecs} and the improvement is 4-6\% for all three metrics. Note that this experimental setting provides a much harder challenge than the earlier binary classification and the results prove that the embeddings learned using {\tt seq2vec} are superior t learned using {\tt ProtVecs} for this task. 
 
\subsection{kNN based Classification}\label{sec:QoE}
In an ideal scenario, distributed representations should embed sequences in such a manner that different families are separated from each other and the sequences belonging to same class are clustered together. This provides us with a way to do classification using a $k$-nearest neighbor based approach. We first embed the training and test data in a vector space using either {\tt seq2vec} or {\tt ProtVecs}. In order to assign a family to a test sequence, we find its $k$-nearest training sequences in the vector space and predict the family using a majority vote. Here we assume that we embed both the train and test data simultaneously. If test data is not available for embedding, we can later use gradient-descent to learn the vector representation of the test sequences as explained in \cite{le2014distributed}.  

\begin{figure}[h]
    \centering
        \includegraphics[width=0.4\textwidth]{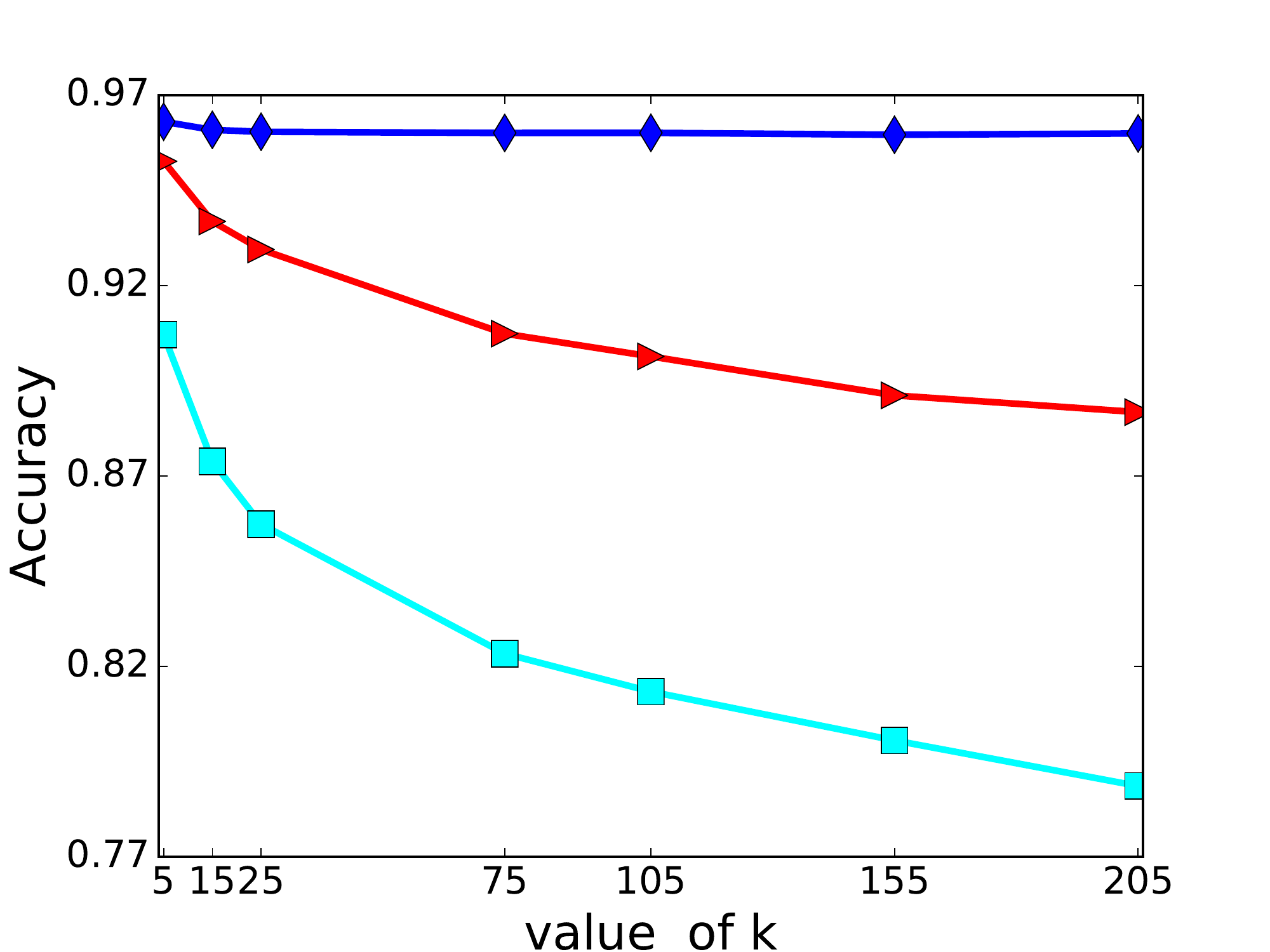}
    \caption{kNN Classification: Performance of \textcolor{red}{{\tt seq2vec}} (\textcolor{red}{--$\blacktriangleright$--$\blacktriangleright$--}) and \textcolor{cyan}{{\tt ProtVecs}} (\textcolor{cyan}{--$\blacksquare$--$\blacksquare$--}) as compared to that of \textcolor{blue}{BLAST} (\textcolor{blue}{--$\blacklozenge$--$\blacklozenge$--}) as a function of $k$ used for kNN.}
    \label{fig:QOE}
\end{figure}

We performed this experiment for the $25$ biggest families consisting of more than $1000$ sequences each, reporting results based on $10$-fold cross validation in Figure \ref{fig:QOE}. It is clear that {\tt seq2vec} based embeddings are significantly better than the ones produced by {\tt ProtVecs}.

\subsubsection{Classification using BLAST}\label{Comp with BLAST}
The BLAST heuristic finds the most similar sequences to a given query sequence based on a hit extension strategy. BLAST ranks matches to a query sequence according to an estimate of the statistical significance of the match, and reports the base identity via a bit score. This requires processing of actual sequences to compute similarity, while we represent these sequences in a low-dimensional space using {\tt seq2vec}. Since BLAST is in some sense a retrieval tool, in order to do classification, we look at the top-$k$ results returned and use majority voting to assign the query to a protein family. This is similar to the setup of kNN classification in Section~\ref{sec:QoE} and hence can be compared directly. The results are reported in Figure \ref{fig:QOE}.

It is clear that while {\tt seq2vec} managed to perform significantly  better than {\tt ProtVecs}, BLAST is markedly superior for this task. We note that the scoring inherent in BLAST closely follows the evolutionary relationships observed across large numbers of protein alignments, and on which protein families are defined.  An important focus for future work is thus to explore the extent to 
which this information can be captured implicitly in the context modelling of our approach, and the performance of embedding based approaches improved to become genuinely competitive with BLAST.

%% file: conclusions.tex
\section{Conclusions and Future Work}
In this paper, we proposed {\tt seq2vec} -- a distributed representation framework for biological sequences. We demonstrated the advantage of our approach by utilizing the learned embeddings for protein classification problems. {\tt seq2vec} performs significantly better than the current state-of-the-art embedding based approach.

Going forward, we would like to learn embeddings for other biological sequences and use them for tasks such as homology detection in nucleotide sequences,  functional annotation, and the prediction of structure and regulatory relationships. As noted in the last section, there is a significant performance gap between {\tt seq2vec} and BLAST for protein classification, and this may perhaps be addressed by embeddings tailored to reflect the similarity weightings inherent in established substitution matrices. Moreover, the embedding approach may allow us to capture important context where substitution matrices or other alignment derived scoring methods are not available, and to improve the state of the art for a range of bioinformatics tasks which presently rely on more complex and costly feature sets. 
